\documentclass[10pt,twocolumn,letterpaper]{article}

\usepackage{wacv}
\usepackage{times}
\usepackage{epsfig}
\usepackage{graphicx}
\usepackage{amsmath}
\usepackage{amssymb}

\usepackage{multirow}
\usepackage{amsmath}
\usepackage{dblfloatfix}

\DeclareMathOperator*{\argmin}{arg\,min}
\usepackage[symbol]{footmisc}

\wacvfinalcopy 

\def\wacvPaperID{****} 

\ifwacvfinal\fi
\begin{document}

\title{Leveraging Filter Correlations for Deep Model Compression}

\author{
Pravendra Singh$^{}$\footnotemark[1] \hspace{1cm}Vinay Kumar Verma$^{}$\footnotemark[1] \hspace{1cm}Piyush Rai\hspace{1cm}Vinay P. Namboodiri\\
Department of Computer Science and Engineering, IIT Kanpur, India\\
{\tt\small \{psingh, vkverma, rpiyush, vinaypn\}@iitk.ac.in}
}

\maketitle
\ifwacvfinal\thispagestyle{empty}\fi

\begin{abstract}
We present a filter correlation based model compression approach for deep convolutional neural networks. Our approach iteratively identifies pairs of filters with the largest pairwise correlations and drops one of the filters from each such pair. However, instead of discarding one of the filters from each such pair na\"{i}vely, the model is re-optimized to make the filters in these pairs maximally correlated, so that discarding one of the filters from the pair results in minimal information loss. Moreover, after discarding the filters in each round, we further finetune the model to recover from the potential small loss incurred by the compression. We evaluate our proposed approach using a comprehensive set of experiments and ablation studies. Our compression method yields state-of-the-art FLOPs compression rates on various benchmarks, such as LeNet-5, VGG-16, and ResNet-50,56, while still achieving excellent predictive performance for tasks such as object detection on benchmark datasets.
\end{abstract}

\vspace{-1em}
\section{Introduction}
\footnotetext[1]{Equal contribution.}

Recent advances in convolutional neural networks (CNN) have yielded state-of-the-art results on many computer vision tasks, such as classification, detection, etc.  However,  training data and storage/computational power requirements limit their usage in many settings. To address this, one line of research in this direction has focused on training CNNs with limited data \cite{finn2018probabilistic,verma2017generalized,kim2018bayesian,reed2018fewshot,qiao2017few,wang2017zero}. Another line of work has focused on model compression to make it more efficient in terms of FLOPs (speedup) and memory requirements. The memory requirement in CNNs can be viewed either as runtime CPU/GPU memory usage or storage space for the model. A number of recent works \cite{denton2014exploiting,abbasi2017structural,singh2019hetconvijcv,singh2019hetconv,singh2019stability,singh2019multi,zhang2015efficient,he2018soft,channelPruning17,singh2019falf,mazumder2019cpwc,singh2019play,singh2020cooperative,singh2019accuracy,thiNet17,li2016pruning} have explored such possibilities for efficient deep learning.

Most existing model compression/pruning methods can be divided into three categories. The first category  \cite{chen2015compressing,han2015deep} has broadly considered introducing sparsity into the model parameters. These approaches give a limited compression rate on FLOPs and Total Runtime Memory (TRM), and need special software (sparse libraries) support to get the desired compression. These approaches provide a good compression rate in terms of weights storage, but provide limited FLOPs and  TRM compression.

The second category of methods \cite{han2015deep,binarycompression,louizos2017bayesian,polino2018model,mishra2017apprentice} has broadly focused on quantization based pruning. Special hardware is needed to provide the acceleration for the final compressed model. These kinds of model compression techniques are primarily designed for IoT devices. 

The third category of methods \cite{denton2014exploiting,abbasi2017structural,zhang2015efficient,singh2019multi,singh2019stability} focuses on filter pruning. These approaches are generic and can be used practically without needing any special software/hardware needed for acceleration. These approaches provide a high compression rate in terms of FLOPs and TRM because of pruning of the whole convolutional filters from the model, which also reduces the depth of the feature maps. Sparsity and quantization based approaches are complementary to these approaches.

Filter pruning approaches require some measure to calculate the importance of the filter, which is a difficult task in general, and many heuristics have been used to measure filter importance. For example, \cite{abbasi2017structural} used a brute-force approach to discard the filters. They prune each filter sequentially and measure the importance of filters based on their corresponding accuracy drop, which can be impractical for large networks. \cite{li2016pruning} uses the $\ell_1$ norm (sum of absolute values) of the filter to measure the filter importance, assuming that a high $\ell_1$ norm filter is more likely to have a bigger influence on the feature map. \cite{molchanov2016pruning,sharma2017incredible} use Taylor expansion based filter importance, which is motivated by the early work on optimal brain damage \cite{lecun1990optimal,hassibi1993second}. 

As discussed in \cite{he2018pruning}, filter importance based pruning methods have certain limitations in the form of requirements that are not always met. Compressed models produced by such methods suffer from redundancy because these methods don't consider filter redundancy while pruning. Therefore, the filter importance based pruning method is unable to reduce the redundancy present in the model and achieve the optimal solution.

In this work, we propose iteratively removing the filters containing redundant information and getting a subset of filters that are minimally correlated. As also evidenced in other recent work, uncorrelated filters help to reduce overfitting \cite{3cogswell2015reducing,4rodriguez2016regularizing} and give a compact model with minimal redundancy. There are no benefits by keeping redundant filters in the model, and it will only result in overfitting or reduced generalization performance \cite{3cogswell2015reducing,4rodriguez2016regularizing}. The works in \cite{ayinde2018building,wang2018exploring} also eliminate the redundancy in convolutional filters by applying clustering to feature maps to perform filter pruning. In contrast, we use filter correlation to measure the redundancy present in the pairs of filters.

In particular, we present a filter pruning approach based on the correlation coefficient (Pearson correlation coefficient) of filter pairs. Unlike other prior works \cite{he2018soft,channelPruning17,thiNet17,li2016pruning} on filter pruning, instead of measuring \emph{individual} filter importance, we measure importance for each \emph{pair} of filters. Filter pairs that have the largest correlation (high redundancy) are given the lowest importance and are chosen for further optimization, before eventual pruning. In the optimization step, we further increase the correlation between each chosen filter pair and then finally prune (discard) one filter from the pair. This optimization step helps to transfer the knowledge of the filter to another one before discarding it. For example, suppose two filters $f_1,f_2$ in a pair have 60\% pairwise correlation. If we discard one filter, we lose some of the information. However, suppose prior to pruning, we optimize the model in a way such that the filter pair's correlation increases to 99\%. If we now discard one of the filters, we lose little information, and it would be safe to prune one of the filters from the pair (since the two filters in the pair have a high degree of similarity), and finetuning can quickly recover it. Our approach starts with the pre-trained model and then iteratively prunes the redundant filters as shown in Figure~\ref{fig:main}. 

\section{Related Work}
Most of the recent work on deep model compression can be categorized into three broad categories.

\subsection{Connection Pruning}
Connection pruning is a direct way to introduce sparsity into the CNN model. One approach for CNN compression is to prune the unimportant parameters. However, it is challenging to define the importance of parameters quantitatively. There are several approaches to measure the importance of the parameters. Optimal Brain Damage \cite{lecun1990optimal} and Optimal Brain Surgeon \cite{hassibi1993second} used the second order Taylor expansion to calculate the parameters importance. However, the second order derivative calculations are very costly. \cite{chen2015compressing} used hashing to randomly group the connection weights into a single bucket and then finetune the network. \cite{wu2018blockdrop} proposed the skip layer approach for network compression. \cite{han2015deep} proposed an iterative approach where absolute values of weights below a certain threshold are set to zero, and the drop in accuracy is recovered by finetuning. This approach is very successful when most of the parameters lie in the fully connected layer. The main limitation of these approaches is the requirement of special hardware/software for acceleration at run-time.

\subsection{Filter Pruning}
Filter pruning approaches (which is the focus of our work too) do not need any special hardware or software for acceleration. The basic idea in filter pruning \cite{he2018soft,channelPruning17,thiNet17,li2016pruning} is to get an estimate of the importance of the filters and discard the unimportant ones. After that, at each pruning step, re-training is needed to recover from the accuracy drop.  \cite{hu2016network} evaluates the importance of filter on a subset of the training data based on the output feature map.  \cite{abbasi2017structural} used a greedy approach for pruning. They evaluated the filter importance by checking the model accuracy after pruning the filter.  \cite{molchanov2016pruning,li2016pruning} used similar approach but different measure for filter pruning. \cite{denton2014exploiting,zhang2015efficient,jaderberg2014speeding} used the low-rank approximation. \cite{singh2019stability} used a data-driven approach for calculating filter importance and pruning.
\cite{liu2017learning} performed the channel level pruning based on the scaling factor in the training process. Recently, group sparsity is also a popular method for filter pruning. \cite{singh2019multi,lebedev2016fast,wen2016learning,zhou2016less,alvarez2016learning} explored the filter pruning based on the group sparsity.

\begin{figure}[t]
    \centering
    \includegraphics[scale=.26]{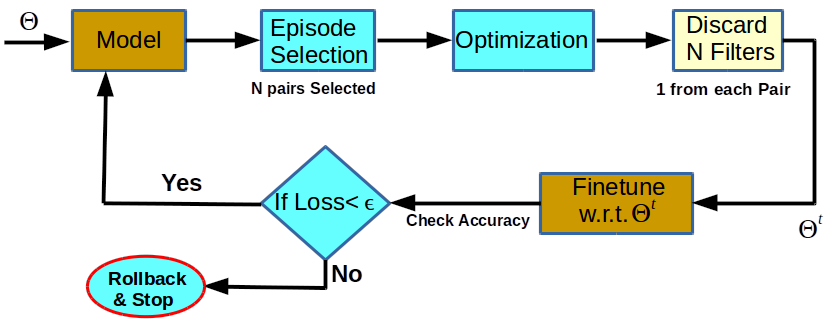}
    \caption{Correlation-based filter pruning where strongly correlated filter pairs are selected for optimization. The compression of the model depends on the user-defined error tolerance limit ($\epsilon$). Hence $\epsilon$ can be seen as the stopping criteria in our approach. }
    \label{fig:main}
    
\end{figure}

\subsection{Quantization}
Weight quantization based approaches have also been used in prior works on model compression. \cite{han2015deep,dubey2018coreset,Tung_2018_CVPR} compressed CNN by combining pruning, quantization, and huffman coding. \cite{miao2017towards} conducted the network compression based on the float value quantization for model storage. Binarization \cite{binarycompression} can be used for the network compression where each float value is quantized to a binary value. Bayesian methods \cite{louizos2017bayesian} are also used for the network quantization. The quantization methods require special hardware support to get the advantage of the compression. 

Our method is generic and does not require any special hardware/software, with the special support we can further increase the FLOPs and memory compression because our method is complementary to other pruning methods such as binary/quantized weights, connection pruning, etc. 

\section{Proposed Approach}
\subsection{Terminology}
Let $\mathcal{L}_i$ be the $i^{th}$ layer and $i\in [1,2,\dots K]$ where $K$ is the number of convolutional layers. The layer  $\mathcal{L}_i$  has $c_o$ filters which is the number of output channels. The set of filters at layer $\mathcal{L}_i$ is denoted as $\mathcal{F}_{\mathcal{L}_i}$, where $\mathcal{F}_{\mathcal{L}_i}=\{f_1,f_2,\dots, f_{c_o}\}$. Each filter $f_i$ is of dimension $(h_k,w_k,c_{in})$, where $h_k$, $w_k$ and $c_{in}$ are height, width and number of input channels, respectively.

\subsection{FLOPs and Memory Size Requirements}
Here we provide a brief analysis to help illustrate the effect of architecture hyperparameters on the FLOPs and memory consumption (which we will use in our experimental results to compare the various approaches).

For a CNN model, the total number of FLOPs on the layer $\mathcal{L}_i$ (convolutional $(FLOPs_{conv})$ or fully connected $(FLOPs_{fc})$) with the batch size B  can be given as:
\begin{equation}\label{flopconv}
    FLOPs_{conv_i}= c_{in}w_k h_k w_o h_o c_{o}*B
\end{equation}
\begin{equation}\label{flopfc}
    FLOPs_{fc_i}= c_{in} c_{o}*B
\end{equation}
Here ($w_{in},h_{in},c_{in}$) is the input feature map, ($w_k,h_k,c_{in}$) is the convolutional filter and ($w_o,h_o,c_o$) is the output feature map. The total FLOPs across network can be defined as:
\begin{equation}
   FLOPs=\sum_{i=1}^{K}  FLOPs_{{conv}_i} +\sum_{j=1}^{N}  FLOPs_{{fc}_j}
\end{equation}
Here $K,N$  is the number of convolutional and fully connected layers respectively. Convolutional filter  of dimension ($w_k,h_k,c_{in}$) is  applied $w_o h_o c_{o}$ times to the input feature map of dimension ($w_{in},h_{in},c_{in}$) to produce output feature map of dimension ($w_o,h_o,c_o$). 
Also, there are two sources of memory consumption: 1- feature map size, 2- parameter weight size. There are some other memory consumptions as well, but these are not feasible to estimate like those that are related to the implementation details of the model, the framework used, etc. So we can estimate the lower bound of memory size. The estimated memory requirement for layer $\mathcal{L}_i$ can be calculated as
\begin{equation}\label{mfeaturemap}
    M_{fm_i}=4w_{o}h_{o}c_{o}*B
\end{equation}
\begin{equation}\label{mweight}
    M_{w_i}= 4w_kh_kc_{in}c_o
\end{equation}
Where $M_{fm_i}$ is memory required for the feature map ($w_o,h_o,c_o$) and $M_{w_i}$ is the memory required for parameter storage  at layer $\mathcal{L}_i$.  So the total memory requirement across each layers can be calculated as:
\begin{equation}
    TRM=\sum_{i=1}^{K+N}  M_{fm_i} +\sum_{j=1}^{K+N}  M_{w_j}
\end{equation}

For the fully connected layer $w_k,h_k,w_o,h_o=1$ and $c_{in},c_o$ are  the number of incoming and outgoing connections respectively. For the convolutional layer $c_{in}, c_{o}$ is the number of input and output channel respectively. Please note that $M_{fm_i}$ depends on the batch size. Therefore the methods that are based on sparsity but not filter pruning only reduce $M_{w_i}$. For such approaches, the feature map size remains the same and grows linearly with respect to the batch size.

\subsection{Our Approach}

Our approach iterative prunes a pre-trained model. In each iteration, we choose correlated filter pairs from each layer and optimize the selected filter pairs such that filters in each selected pair are as highly correlated as possible. This enables us to safely discard, without information loss, one of the filters (which is redundant) from each of the selected pairs. We finetune the compressed model after pruning. This constitutes one iteration of pruning (Figure~\ref{fig:main}), which can be further repeated to more iterations until we could recover accuracy with in the tolerance limit.

In our approach, we consider the importance of filter \emph{pairs} for pruning. We use the correlation coefficient of the filter pair to quantify their importance. The filter pair that has the largest correlation coefficient is defined as the \emph{least important}. It is considered as the least important filter pair because of the presence of redundancy (one of the filters from the pair is redundant). However, if we drop one of the filters from this pair, we might also be losing their mutually-complementary information, which may or may not be captured by the other filters. Therefore, before discarding one of the filters from the pair, we need to transfer this information to the other filters by optimizing the model before pruning. There may be the case when a filter belongs to multiple highly correlated pairs, then that filter is pruned in the process. It may also be possible that both filters are important in highly correlated filter pair, and accuracy may drop after pruning, but the small finetuning (1-2 epoch) will recover it as we have only removed a redundant filter from the pair.

\begin{figure}[t]
    \centering
    \includegraphics[height=5cm, width=8cm]{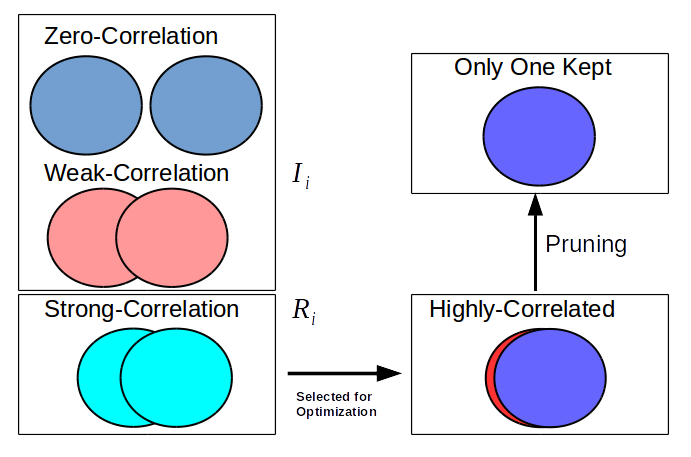}
    \caption{Strongly correlated filters are selected and make them highly correlated. Finally, one redundant filter is pruned.}
    \label{fig:filter_selection}
    
\end{figure}

Some of the previous approaches like \cite{ding2018auto,han2015deep,li2016pruning} used the $\ell_1$ or $\ell_2$ regularizer for defining the filter importance. However, using the $\ell_1$ or $\ell_2$ regularizer typically works well only when there is significant redundancy in the model. These approaches are unable to give a highly compressed model without sacrificing accuracy. If we try to get a highly compact model with these approaches, the system performance degrades rapidly because the strong $\ell_1$ or $\ell_2$ regularizer starts penalizing the important filters as well and, therefore, cannot achieve a significant compression. If we increase the $\ell_2$ penalty, it starts controlling the model complexity. Large $\ell_2$ penalty results in underfitting. Therefore these approaches can only lead to limited levels of filter pruning. In contrast, in our proposed approach, we learn a minimal basis of filters that are highly uncorrelated. Figure~\ref{fig:filter_selection} illustrates the basic idea of our approach.

Correlation is the most common and one of the simplest statistical measures to quantify data dependence or association. Correlation often refers to how close two variables are to have a linear relationship with each other. Let \textbf{X} and \textbf{Y} are two random variables with expected values $\mu_X$ and $\mu_Y$ and standard deviations $\sigma_X$ and $\sigma_Y$ respectively. The correlation $\rho_{XY}$ can be defined as
\begin{equation}\label{eq:corr}
    \rho_{XY}=\frac{\mathbb{E}[(\mathbf{X}-\mu _{X})(\mathbf{Y}-\mu _{Y})]}{\sigma _{X}\sigma _{Y}}
\end{equation}Here $ \rho _{XY} \in [-1,1]$, with a high negative or high positive value indicates a high degree of linear dependence between the two variables. When the correlation is near zero, the two variables are linearly independent.

\subsubsection{Episode Selection}
Let us define the two terms which we will use in the proposed pruning process - ``Ready-to-prune ($\mathbf{R_i}$)" and ``Pruned ($\mathbf{P_i}$)". Here $\mathbf{R_i}$ denotes filter pairs selected from the layer $\mathcal{L}_i$ for the optimization process (details in the next section).  $\mathbf{P_i}$ denotes the filters that are eventually selected (one from each pair of $\mathbf{R_i}$) for pruning from the model after optimization. Therefore if $N$ filter pairs are selected in $\mathbf{R_i}$ then $|\mathbf{P_i}|=N$, i.e., from the layer $\mathcal{L}_i$, $N$ filters will get pruned.

In each layer $\mathcal{L}_i$, we find out the filter pairs that have the maximum correlation. For calculating the correlation, we select all filters $\mathcal{F}_{\mathcal{L}_i}=\{f_1,f_2,\dots, f_{c_o}\}$ on a layer $\mathcal{L}_i$ with $c_o$ output channels. Each filter ${f}_i$ on the layer $\mathcal{L}_i$ is  of dimension ($w_{k},h_{k},c_{in}$) is flattened  to a vector of size $w_k \times h_k \times c_{in}$. Now we can calculate the correlation coefficient for filter pair by using Eq.~\ref{eq:corr}. In our approach, we have considered the magnitude of the correlation coefficient, thereby giving the same importance for positive and negative correlation values. Based on the magnitude of the correlation coefficient of each pair, filter pairs are ordered.

The filter pair with the largest correlation value has the minimum importance. Some least important filter pairs are selected for the ready-to-prune set $\mathbf{R_i}$ (select the filter pair ($f_a,f_b$) such that $a\neq b$).    
Let $\mathbf{I_i}$ are the remaining filter pairs at layer $\mathcal{L}_i$. Then:
\begin{equation}
    \mathcal{PF}_{\mathcal{L}_i}=\mathbf{R_i} \cup \mathbf{I_i} \quad s.t. \quad \mathbf{R_i} \cap \mathbf{I_i}=\phi
\end{equation}
Here $\mathcal{PF}_{\mathcal{L}_i}=\{(f_a,f_b) : a \in \{1,2,\dots c_o\}, \quad b \in \{1,2,\dots c_o\}\}$.
The same process is repeated for each layer. This selected set of filter pairs from all the layers ($\mathbf{S_t}$) is called one episode. Where $\mathbf{S_t}$ is the set of filter pairs selected at  $t^{th}$ episode from all $K$ convolutional layers.
\begin{equation}
    \mathbf{S_t}= \{\mathbf{R_1}, \mathbf{R_2}, \dots, \mathbf{R_K}\}
\end{equation}
This set $\mathbf{S_t}$ is the collection of the ready-to-prune filter pairs from all the layers and used for further optimization such that both filters in each pair contain similar information after optimization. Therefore we can safely remove one filter from each pair. The optimization process is explained in the next section.

\subsubsection{Optimization}

Let $C(\Theta)$ be the cost function of the original model and $\Theta$ be the model  parameters. We optimize this cost function with the new regularizer applied to the selected episode (set of filter pairs) $\mathbf{S_t}$.  Our new regularizer $(C_{\mathbf{S_t}})$ is as follows:
\begin{equation}\label{eq:corrregularizer}
    C_{\mathbf{S_t}}=\exp\left(- \sum_{X,Y \in \mathbf{R_i}, \\ \forall \mathbf{R_i} \in \mathbf{S_t}} |\rho_{XY}| \right)
\end{equation}

Here $|\rho_{XY}|$ is the magnitude of correlation coefficient of the filter pair $(X,Y)$ in $\mathbf{R_i}$ and $\mathbf{R_i}\in \mathbf{S_t}$. The idea is here to make the strongly correlated filter pair  highly correlated (as illustrated in Figure~\ref{fig:filter_selection}). Note that Eq.~\ref{eq:corrregularizer} will be minimized when $\sum_{X,Y \in \mathbf{R_i}, \forall \mathbf{R_i} \in \mathbf{S_t}} |\rho_{XY}|$ term is maximum, i.e. each filter pair's magnitude of correlation coefficient $|\rho_{XY}|\to 1$. This new regularizer is added to the original cost function so our new objective function is as follows:
\begin{equation}\label{eq:mainloss}
    \Theta=\argmin_{\Theta}\left( C(\Theta) + \lambda*C_{\mathbf{S_t}} \right)
\end{equation}
Here $\lambda$ is the hyper-parameter to control the regularization term. Minimizing the Eq.~\ref{eq:mainloss} will optimize such that without degrading the model performance, it increases the correlation as high as possible between filters in each pair belonging to $\mathbf{R_i}$. Therefore we can safely remove one filter from each pair. 

\subsubsection{Pruning and Finetuning}
After increasing the correlation between filters in each pair belonging to $\mathbf{R_i}$, we can prune one filter from each pair belonging to $\mathbf{R_i}$.
Our model has a reduced set of the parameter $\Theta'$
\begin{equation}
    \Theta'=\Theta \setminus  \{ \mathbf{p_1},\mathbf{p_2} \dots \mathbf{p_k}\}
\end{equation}
Where $\mathbf{p_i}$ is the set of filters finally selected to be removed from the model. 

Further, we finetune the model w.r.t. the parameter $\Theta'$. Since we discarded the \emph{redundant} filters from the model, the information loss from the model would be minimum. Hence the finetuning process can easily recover from the small loss, and the model's performance can be brought back to be nearly the same as the original one. Please note that, if two filters are highly correlated, then their output feature maps are also highly correlated because both filters are applied to the same input feature maps.

\section{Experiments}
To evaluate our approach Correlated Filter Pruning (CFP), we use three standard models, LeNet-5 \cite{lecun1998gradient}, VGG-16 \cite{vgg2014very} and ResNet-50,56 \cite{resnet}, for classification, and two popular models, Faster-RCNN and SSD, for object detection. Our experiments were done on GTX 1080 Ti GPU and i7-4770 CPU@3.40GHz. Through an extensive set of experiments, we show that our proposed approach achieves state-of-art compression results. In all our experiments, we set $\lambda = 1$ to enforce a high correlation for optimizing Eq.~\ref{eq:mainloss}. We can also choose a smaller value of $\lambda$, but then it would increase the number of epochs in the optimization step. A very high value of $\lambda$ may result in accuracy loss because now optimization gives more weight to $C_{\mathbf{S_t}}$ (Eq.~\ref{eq:mainloss}) than $C(\Theta)$. We empirically found $\lambda = 1$ is the right choice for all our experiments. Filter pairs selection for the optimization process ($\mathbf{R_i}$) is proportional to the FLOPs on layer $\mathcal{L}_i$ to reduce the same \% of FLOPs from every layer. We simultaneously prune filters across all layers. We continued this filter pairs selection strategy until we can recover the accuracy with in the tolerance limit. Ones the tolerance limit is achieved, we start individual layer pruning.  We pruned each layer sequentially from start to end one by one until we could recover accuracy with in the tolerance limit. 

\begin{table}[t]
    \centering
    \scalebox{.82}{
    \addtolength{\tabcolsep}{-2.5pt}
    \begin{tabular}{|l| c| c| c| c|} 
    \hline
    Method & $r_1,r_2$ & Error\% & FLOPs & Pruned Flop \% \\ [0.8ex] 
    \hline\hline
     
    SSL-2 \cite{wen2016learning} & 5,19 & 0.80 & $5.97\times 10^5$ & 86.42\\
    SSL-3 \cite{wen2016learning} & 3,12 & 1.00 & $2.89\times 10^5$ & 93.42\\
    SBP \cite{neklyudov2017structured} & -- & 0.86 & -- & 90.47\\
    SparseVD \cite{molchanov2017variational} & -- & 0.75 & -- & 54.34\\
    \hline\hline
   Baseline & 20,50 & $0.83\pm0.09$ & $4.40 \times 10^6$ & 0.0\\
    \textbf{CFP-1 (Ours)} &\textbf{4,5} & $\textbf{0.91}\pm\textbf{0.07}$& $\mathbf{1.95 \times 10^5}$ & \textbf{95.56}\\
     \textbf{CFP-2 (Ours)} &\textbf{3,5} &  $\textbf{0.95}\pm\textbf{0.05}$& $\mathbf{1.58 \times 10^5}$ & \textbf{96.41}\\
     \textbf{CFP-3 (Ours)} &\textbf{3,4} &   $\textbf{1.20}\pm\textbf{0.09}$& $\mathbf{1.39 \times 10^5}$ & \textbf{96.84}\\
     \textbf{CFP-4 (Ours)} &\textbf{2,3} &   $\textbf{1.77}\pm\textbf{0.08}$& $\mathbf{0.89 \times 10^5}$ & \textbf{97.98}\\
    \hline
    \end{tabular}}
    \vspace{2pt}
    \caption{Pruning results for the LeNet-5 architecture on MNIST. Here $r_1,r_2$ are used to denote the number of remaining filters in first and second convolutional layers respectively. We run each experiment three times and report the ``mean$\pm$std".}
    \label{tab:lenet5}
\end{table}

\subsection{LeNet-5 on MNIST}
MNIST is a handwritten digit dataset contains 60,000 images for training and 10,000 images for the testing. We use the LeNet-5 architecture that contains two convolutional layers and two fully connected layers. The complete architecture is 20-50-800-500, where 20, 50 are the number of convolutional filters in first and second convolutional layers respectively. We trained the model from scratch and achieved an error rate of 0.83\%. 

To show the effectiveness of our proposed approach, we conduct the first experiment on LeNet-5 for the MNIST dataset. As compared to the previous approaches, we achieve a much higher FLOPs compression rate with a relatively small accuracy drop. In prior work, SSL-3 \cite{wen2016learning}, report an error of 1.0\% on 93.42\% FLOPs pruning while we achieve 96.41\% FLOPs pruning with only 0.95\% error. Also, note that, as compared to SBP \cite{neklyudov2017structured}, that has only 90.47\% FLOPs pruning with 0.86\% error, our approach has 95.56\% FLOPs pruning with the negligible (0.05\%) error difference. Please refer to Table~\ref{tab:lenet5} for a detailed comparison. CFP-1 denotes the first compressed model in this iterative pruning scheme when it shows a competitive accuracy as compared to other approaches. We can repeat the process for more iterations to compress the model further and can monitor the accuracy after each pruning iteration. This helps us assess our approach's ability to compress the model further. CPF-2,3 and 4 denote the compressed models obtained by such iterative pruning after 2, 3, and 4 iterations, respectively.

\subsection{VGG-16 on CIFAR-10}
\begin{figure}[t]
    \centering
    \includegraphics[scale=0.32]{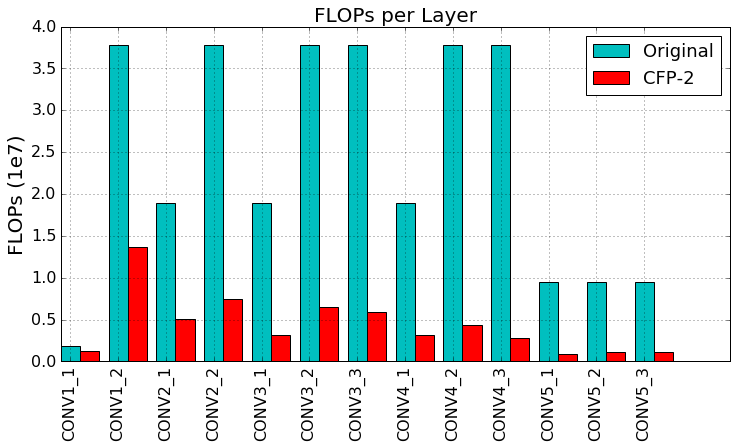}
    \caption{The original and pruned model FLOPs on each layer for VGG-16 on CIFAR-10.}
    \label{fig:channel-flops}
\end{figure}

We next experiment with the VGG-16 architecture on CIFAR-10. Each image size is of size $32\times32$ on the RGB scale. The VGG-16 convolutional layers architecture is the same as \cite{vgg2014very}, and after each convolutional layer, batch normalization layers are added. We use the same architecture and settings as described in \cite{li2016pruning}. The network is trained from scratch and achieves a 6.51\% error rate. Figure~\ref{fig:channel-flops} shows the layer-wise FLOPs distribution for the original and pruned model. 

Like the LeNet-5 pruning results, we observe the same pattern for the VGG-16 pruning on the CIFAR-10 dataset. We have 80.36\% FLOPs pruning with a 6.77\% error while previous state-of-art approach SBPa \cite{neklyudov2017structured} has only 68.35\% FLOPs pruning with 9.0\% error. SparseVD \cite{molchanov2017variational} has 55.95\% pruning with 7.2\% error, while we have 81.93\% pruning with  significantly less (7.02\%) error. Please refer to Table~\ref{tab:vgg-16} for detail comparison. In the Table~\ref{tab:vgg-16}, CFP-1, and CPF-2 denote the first and second compressed models respectively (in this iterative pruning scheme). The original and pruned FLOPs are shown in Figure~\ref{fig:channel-flops}. 

\begin{table}[t]
    \centering
    \scalebox{0.85}{
    \begin{tabular}{|l| c| c| c|} 
    \hline
    Method &  Error\% & FLOPs & Pruned Flop\% \\ [0.8ex] 
    \hline\hline
    Li-pruned  \cite{li2016pruning} &  6.60 & $2.06\times 10^8$ &34.20\\
    SBPa \cite{neklyudov2017structured} &  9.00 & -- & 68.35\\
    SBP  \cite{neklyudov2017structured} &  7.50 & -- & 56.52\\
    SparseVD \cite{molchanov2017variational} &  7.20 & -- & 55.95\\
    \hline\hline
    Baseline  & $6.51\pm0.23$ & $3.137 \times 10^8$ & 0.0\\
    \textbf{CFP-1 (Ours)} &  $\textbf{6.77}\pm\textbf{0.19}$&  $\mathbf{6.16\times10^7}$ & \textbf{80.36} \\
    \textbf{CFP-2 (Ours)} & $\textbf{7.02}\pm\textbf{0.16}$&  $\mathbf{5.67\times10^7}$ & \textbf{81.93} \\
    \hline
    \end{tabular}
    }
    \vspace{2pt}
    \caption{Pruning results for VGG-16 architecture on the CIFAR-10 dataset. We run each experiment three times and report the ``mean$\pm$std".}
    \label{tab:vgg-16}
\end{table}

\subsection{Results on ResNet}
\subsubsection{ResNet-56 on CIFAR-10}
We use the ResNet-56 architecture \cite{resnet} on the CIFAR-10 dataset, which contains the three stages of the convolutional layer of size 16-32-64, where each convolution layer in each stage contains the same 2.36M FLOPs. We trained the model from scratch using the same parameters proposed by \cite{resnet} and achieve the error rate of 6.43\%. 
The network is pruned into two cycles. In the initial cycle, we selected  1 filter pair from each convolutional layer in the first stage (total 9 filter pairs), 2 filter pairs from each convolutional layer in the second stage (total 18 filter pairs) and 4 filter pairs from each convolutional layer in the third stage (total 36 filter pairs) to prune the same amount of FLOPs from each stage for $\mathbf{S_t}$ and pruned one filter from each pair (total 9, 18 and 36 filters pruned from first, second and third stage respectively), till we can recover the accuracy drop with the $\epsilon$ tolerance. In the second cycle, we selected one filter pair (total nine filter pairs from all convolutional layers in a particular stage) in a particular stage for the $\mathbf{S_t}$ and pruning continue till we can recover the model accuracy with the $\epsilon$ tolerance. The results are shown in Table~\ref{tab:resnet-56}. It is clear from the table that as compared to the previous approach, our method shows the state of art result. CP \cite{channelPruning17} has 50.0\% FLOPs pruning with a 8.2\% error, while with only 7.37\% error, we have significantly higher 76.56\% FLOPs pruning.
Similarly, SFP \cite{he2018soft} has 52.6\% pruning with a 6.65\% error, and our model has 61.51\% pruning with the same error. Here $r_1,r_2, r_3$ are used to denote the number of remaining filters in each convolutional layer within the three stages. We use the same approach as \cite{ding2018auto} to resolve the skip connection inconsistency during the filter pruning.

\begin{table}[t]
    \centering
    \scalebox{0.85}{
    \addtolength{\tabcolsep}{-2.5pt}
    \begin{tabular}{|l| c| c| c| c|} 
    \hline
    Method & $r_1,r_2,r_3$ & Error\% & FLOPs & Pruned Flop \% \\ [0.8ex] 
    \hline\hline
    Li-A \cite{li2016pruning} & --- & 6.90 & $1.12\times 10^8$ &10.40\\
    Li-B \cite{li2016pruning}& --- & 6.94 & $9.04\times 10^7$ & 27.60\\
    
    NISP \cite{yu2017nisp} & --- & 6.99 & -- & 43.61\\
    CP \cite{channelPruning17} & --- & 8.20 & -- & 50.00\\
    SFP \cite{he2018soft}& --- & 6.65 & -- & 52.60\\
    AMC \cite{He_2018_ECCV}& --- & 8.10 & -- & 50.00\\
    \hline\hline
    Baseline &16,32,64 & $6.43\pm0.15$ & $1.26 \times 10^8$ & 0.0\\
    \textbf{CFP-1} &10,20,38 & $\textbf{6.68}\pm\textbf{0.12}$ & $\mathbf{4.85\times 10^7}$&\textbf{61.51} \\
    \textbf{CFP-2} &9,18,36 & $\textbf{6.93}\pm\textbf{0.10}$ & $\mathbf{4.08\times 10^7}$& \textbf{67.62}\\
    \textbf{CFP-3} &8,16,27 & $\textbf{7.37}\pm\textbf{0.17}$ &$\mathbf{2.95\times 10^7}$ & \textbf{76.59}\\
    \hline
    \end{tabular}
    }
    \vspace{0pt}
    \caption{Pruning results for ResNet-56 architecture on CIFAR-10. We run each experiment three times and report the ``mean$\pm$std".}
    \label{tab:resnet-56}
\end{table}

\subsubsection{ResNet-50 on ImageNet}
We experiment with the ResNet-50 model on the large-scale ImageNet \cite{imagenet2015} dataset. The results are shown in Table~\ref{tab:resnet-50}. We follow the same settings as mentioned in \cite{channelPruning17}. As compared to ThiNet-50, we have similar FLOPs pruning with significantly better accuracy. 

Other proposed approaches, such as channel pruning (CP) \cite{channelPruning17} and structured probabilistic pruning (SPP) \cite{wang2017structured} have $\sim 50\%$ FLOPs pruning, but their error rate is high. In particular, CP has a $9.2\%$ error, and SPP has a $9.6\%$ error. Our proposed approach is highly competitive with these approaches in terms of FLOPs pruning, while also yielding significantly better accuracy. Please refer to Table~\ref{tab:resnet-50} for more details.

\subsection{Ablation Study}

In the following section, we present an ablation study on how our correlation-based criterion helps in preserving the information in the model during filter pruning, how finetuning helps after discarding one of the filters from the filter pair and analyze the correlations among the filters retained in the final model.

\begin{table}[!t]
    \centering
    \scalebox{0.92}{
    \addtolength{\tabcolsep}{8pt}
    \begin{tabular}{|l| c| c| c|} 
    \hline
    Method   & Error (\%)  &  Pruned Flop (\%) \\ [0.8ex] 
    \hline\hline
    
    ThiNet-50 \cite{luo2018thinettp}  & 9.0 & $\sim$ 50\\
    CP \cite{channelPruning17}  & 9.2 & $\sim$ 50\\
    SPP \cite{wang2017structured}  & 9.6 & $\sim$ 50\\
    WAE \cite{chen2017learning}  & 9.6 & 46.8\\
   
   \hline\hline
   Baseline  & 7.8 & 0.0\\
    \textbf{CFP (Ours)}  & \textbf{8.6} & \textbf{49.6}\\
    \hline
    \end{tabular}
    }
    \caption{ResNet-50 Pruning results on the ImageNet with the other state-of-art approaches. The baseline network's top-5 accuracy is 92.2\% (https://github.com/KaimingHe/deep-residual-networks).}
    \vspace{-5pt}
    \label{tab:resnet-50}
\end{table}

\begin{table}
\begin{center}
\scalebox{0.92}{
 \addtolength{\tabcolsep}{5.5pt}
\begin{tabular}{|c|c|c|c|}
\hline 
Model &  OPT & Error (\%) & PF (\%)\\
\hline
\multirow{2}{*}{{LeNet-5 (CFP-3)}} & No & 1.81 & 96.84 \\
 & Yes & \textbf{1.20} & 96.84 \\\hline
\multirow{2}{*}{{LeNet-5 (CFP-4)}} & No & 2.32 & 97.98 \\
 & Yes & \textbf{1.77} & 97.98 \\\hline
 \multirow{2}{*}{{VGG-16 (CFP-1)}} & No & 7.20 & 80.36 \\
 & Yes & \textbf{6.77} & 80.36\\\hline
 \multirow{2}{*}{{VGG-16 (CFP-2)}} & No & 7.61 & 81.93 \\
 & Yes & \textbf{7.02} & 81.93 \\\hline

\end{tabular} 
}
  \caption{Effect of the correlation optimization given by equation-[\ref{eq:corrregularizer}] for the LeNet-5 and VGG-16 (\textbf{OPT}: Optimization, and \textbf{PF}: Pruned Flop).}
  \vspace{-5pt}
    \label{fig:optimization}
\end{center}
\end{table}


\subsubsection{Optimization w.r.t. Accuracy}
Recall that, before discarding filters directly based on correlation, we further optimize the filter correlation such that the strongly correlated filter becomes even more strongly correlated (using the regularizer based on Eq.~\ref{eq:corrregularizer}). We have found that if we discard the filter without optimization, the model suffers from the significant accuracy drop due to the loss of potentially mutually-complementary information. Therefore, before discarding one of the filters from the pair, we need to transfer this information to the other filters that remain in the model after pruning using Eq.~\ref{eq:mainloss}.

Please refer to Table~\ref{fig:optimization} to see the effect of the optimization. LeNet and VGG-16 compressed models achieve better accuracy with optimization.


\subsubsection{Analysis of correlation among filters in the final compressed model}

In the case of VGG-16 on the CIFAR-10, initially, the maximum filter correlation is 0.7, but in the final compressed model, the maximum filter correlation is nearly 0.1, which shows that we have successfully removed the redundant filters. At the same time, the fact that the classification accuracy does not drop much (Table~\ref{tab:vgg-16}), indicates that the useful discriminative filters are preserved. Here, we find a subset of filters that are minimally correlated but preserves maximal information for our finally compressed model. These uncorrelated filters help to reduce the overfitting \cite{3cogswell2015reducing,4rodriguez2016regularizing} and give a compact model with the least possible redundancy.

\subsection{Speedup and Memory Size}
\begin{figure}[!t]
    \centering
    \includegraphics[scale=0.32]{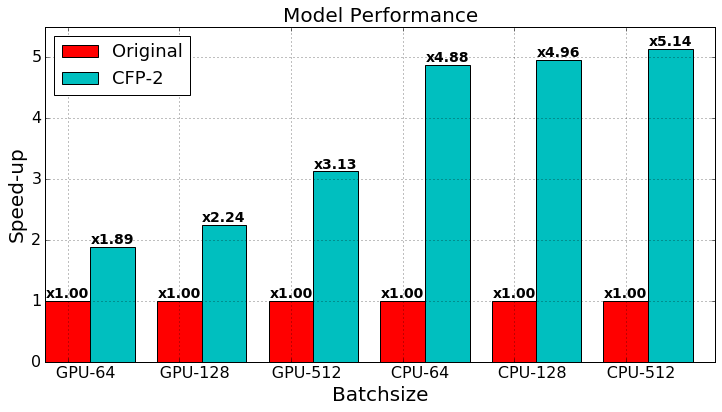}
    \caption{Model performance on the CPU (i7-4770 CPU@3.40GHz) and GPU (TITAN GTX-1080 Ti) for VGG-16 on CIFAR-10. }
    \vspace{-5pt}
    \label{fig:model_performance}
\end{figure}

\begin{figure}[!t]
    \centering
    \includegraphics[scale=0.31]{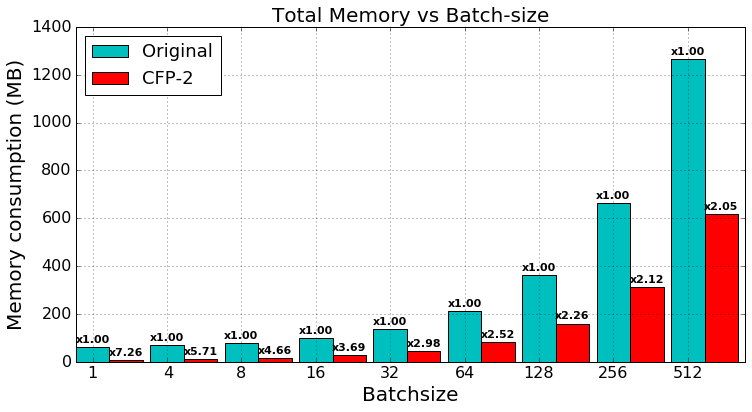}
    \caption{Increase in the total memory size w.r.t. batch size for VGG-16 on CIFAR-10. }
    \vspace{-5pt}
    \label{fig:memory_size}
\end{figure}

The pruned FLOPs is not necessarily equivalent to practical model speedup because FLOPs give the theoretical speedup. The practical speedup can be very different from the result reported in terms of pruned FLOPs percentage. The practical speedup depends on the many other factors, for example, parallelization bottleneck on intermediate layers, I/O operation, etc. Also, total run-time memory (TRM) does not depend only on the compressed model parameters size but also on the feature maps (FM), batch-size (BS), the dynamic library used by Cuda, all the supporting header-file, etc. Here we don`t have control over all the parameters but Model parameters size (MPS), FM, and BS. To show the practical speedup and Memory size, we experiment with the VGG-16 model over the CIFAR-10 dataset. The result for the speedup and TRM are shown in the Figure~\ref{fig:model_performance} and Figure~\ref{fig:memory_size} respectively.

\begin{table*}[t]
\centering
\scalebox{1.0}{
\addtolength{\tabcolsep}{6pt}
\begin{tabular}{|c|c|c|c|c|c|c|}
\hline
\multirow{2}{*}{\textbf{Model}} & \multicolumn{4}{c|}{\textbf{AP}} & \multirow{2}{*}{\textbf{Size}} & \multirow{2}{*}{\textbf{Parameters}} \\ \cline{2-5}
 & \textbf{prohibitory} & \textbf{mandatory} & \textbf{danger} & \textbf{mAP} &  &  \\ \hline
\textbf{SSD512-O} & 96.83 & 86.93 & 87.05 & 90.27 & 98.7 MB & 24.7M \\ \hline
\textbf{SSD512-P} & 98.35 & 88.45 & 89.01 & \textbf{91.94} & \textbf{4.0 MB (24.7$\times$)} & \textbf{0.99M (4.0\%)} \\ \hline
\end{tabular}
}
\vspace{2pt}
\caption{Class wise AP for SSD512-original(O) and SSD512-pruned(P) model on GTSDB dataset.}
\label{tabssdgtsdb}
\vspace{-5pt}
\end{table*}

\begin{table*}[!ht]
\centering
\scalebox{.92}{
\begin{tabular}{|c|c|c|c|c|c|c|c|c|c|c|c|c|c|}
\hline
\multirow{2}{*}{\textbf{Model}}& \multirow{2}{*}{\textbf{data}} & \multicolumn{3}{c|}{\textbf{Avg. Precision, IoU:}}  & \multicolumn{3}{c|}{\textbf{Avg. Precision, Area }} & \multicolumn{3}{c|}{\textbf{Avg. Recall, \#Dets:}} & \multicolumn{3}{c|}{\textbf{Avg. Recall, Area:}}\\ \cline{3-14}
 & & \textbf{0.5:0.95} & \textbf{0.5} & \textbf{0.75}  & \textbf{S} & \textbf{M} & \textbf{L} & \textbf{1} & \textbf{10} & \textbf{100} & \textbf{S} & \textbf{M} & \textbf{L} \\ \hline

\textbf{F-RCNN original} & trainval35K & 27.5 & 48.7 & 28.1  & 12.4 & 31.5 & 38.7  & 25.7 & 38.8 & 39.8 & 21.2 & 45.5 & 55.0 \\ \hline
\textbf{F-RCNN pruned} & trainval35K & 27.8 & 48.4 & 28.5  & 13.3 & 31.6 & 38.6 & 26.1 & 39.5 & 40.6 & 22.6 & 45.7 & 55.4\\ \hline
\end{tabular}
}
\caption{Generalization results are shown on the MS-COCO dataset. Pruned ResNet-50 (CFP) used as a base model for Faster-RCNN. Where S, M, and L are the small, medium, and large area ranges respectively for evaluation. 1, 10, and 100 denotes thresholds on max detections per image (http://cocodataset.org/\#detection-eval).}
\vspace{-5pt}
\label{coco}
\end{table*}

FM is the most important factor for reducing the runtime memory since it grows linearly w.r.t. batch size and quadratic w.r.t. image size while MPS is fixed. The filter pruning approach reduces the model parameters as well as feature maps memory size while all the approaches based on sparsity in the model are reducing only the MPS, and the size of the FM remains the same. Hence batch size has the bottleneck. If we have a limited batch size, this reduces the parallelism in the GPU, resulting in the speed drop. TRM can be defined as:
\begin{equation}
\small
    TRM= MPS+ (FM*4*BS)
\end{equation}
It is clear from Figure~\ref{fig:memory_size} that with the increase in BS, TRM memory increases. Therefore we cannot go for a big batch size. While Figure~\ref{fig:model_performance} shows that for the small batch size system performance (speedup) degraded. Therefore for the speedup, we have to choose a bigger BS, but there is a memory bottleneck on the GPU or CPU. Hence in the proposed approach, we prune the whole convolutional filter so that FM memory can be reduced.

\subsection{Generalization Ability}

\subsubsection{Compression for Object Detection}
To show the generalization ability of our approach, we also show the result on the detection network. In this experiment we have taken two most popular object detector SSD \cite{liu2016ssd} on GTSDB dataset and Faster RCNN \cite{ren2015fasterrcnn} on MS-COCO \cite{lin2014coco}. In the case of SSD, we achieve $\sim$25$\times$ compression in terms of model parameters with significant improvement in AP. For faster RCNN, we have used ResNet-50 as a base network. 

\vspace{-5pt}
\subsubsection{SSD512 on German traffic detection benchmarks}
In this experiment, we evaluate the generalization ability of our pruned model, VGG-16 CFP-2,  which is pruned on CIFAR-10. First, we trained original SSD512 on German traffic detection benchmarks (GTSDB) \cite{Houbengermantrafic} dataset, In which ImageNet pre-trained base network was used. In the second case, we replace the base network of SSD512 with our pruned model, VGG-16 CFP-2. After training, we analyzed the model and found the model is over-fitted because GTSDB is a small scale dataset. Our pruned SSD512 model detects the object from the initial layer only, which is CONV4\_3, and we removed all other remaining layers after CONV4\_3. After doing this, we observed a significant improvement in the mAP and $\sim$25$\times$ compression in model size. Hence our pruned model successfully generalizes the object detection task on the GTSDB dataset. Refer to Table~\ref{tabssdgtsdb} for the detailed experimental results.
\subsubsection{Faster RCNN on COCO}
We experiment on the COCO detection
datasets with 80 object classes \cite{lin2014coco}. 80k
train images and  35k val images are used for training (trainval35K) \cite{lin2017feature}. We report the detection accuracies over the 5k unused val images (minival). In this first, we trained Faster-RCNN with the ImageNet pre-trained ResNet-50 base model. The results are shown in Table~\ref{coco}. 
In the second experiment, we used our pruned ResNet-50 model (CFP), which is pruned on the ILSVRC-2012 dataset as given in Table~\ref{tab:resnet-50}. Then we used our pruned ResNet-50 (CFP) model as a base network in Faster-RCNN. In the Faster-RCNN implementation, we used ROI Align instead of ROI Pooling. We found that the pruned model shows slightly better performances in some cases (mAP@0.75, mAP@0.5:0.95). Refer to the Table~\ref{coco} for more details.
\section{Conclusion}
We have proposed a novel approach for filter pruning, which is guided by pairwise correlations of filters. Unlike the previous heuristics for measuring individual filters importance for pruning, we proposed a new approach for considering filter pairs importance based on the redundancy present in the pair. In the pruning process, we iteratively reduce the redundancy in the model. Our approach, as compared to the existing methods, shows state-of-art results. The efficacy of our method is demonstrated via a comprehensive set of experiments and ablation studies. We have shown the generalization capability of our approach for the object detection task. 

\section*{Acknowledgment:}
PS is supported by the Research-I Foundation at IIT Kanpur. VKV acknowledges support from Visvesvaraya PhD Fellowship and PR acknowledges support from Visvesvaraya Young Faculty Fellowship.  

{\small
\bibliographystyle{ieee}
\bibliography{egbib}
}

\end{document}